\documentclass[journal]{IEEEtran}
\usepackage{amsmath,amsfonts}
\usepackage{algorithmic}
\usepackage{array}
\usepackage[caption=false,font=normalsize,labelfont=sf,textfont=sf]{subfig}
\usepackage{textcomp}
\usepackage{stfloats}
\usepackage{url}
\usepackage{verbatim}
\usepackage{graphicx}

\usepackage[table,xcdraw]{xcolor}
\usepackage{hyperref} 
\usepackage{booktabs}
\usepackage{multirow}

\def\BibTeX{{\rm B\kern-.05em{\sc i\kern-.025em b}\kern-.08em
    T\kern-.1667em\lower.7ex\hbox{E}\kern-.125emX}}
\usepackage{balance}
\begin{document}
\title{Edge-Efficient Transformer for End-to-End RF Spectrum Monitoring}


\author{Zhifan Song, Haralampos-G. Stratigopoulos, and Hassan Aboushady\\
\small{Sorbonne Université, CNRS, LIP6, Paris, France}\vspace{-0.4cm}

\thanks{This work was funded by the Chips JU project Resilient Trust of the EU’s
Horizon Europe research and innovation programme under Grant agreement
No 101112282.}}




\maketitle

\begin{abstract}
We present \emph{E-SpecFormer} (Edge Spectrum monitoring Transformer) for end-to-end automatic modulation and covert channel (CC) recognition. We introduce \emph{LiTAN} (Linear Tanh Attention Network), a Softmax- and LayerNorm-free attention mechanism that reduces complexity while increasing accuracy in RF tasks. E-SpecFormer is parameterized in four scalable variants (Nano, Small, Medium, Large) to accommodate diverse hardware constraints. Using the RadioML2018 dataset for modulation recognition, the Nano variant achieves 86.5\% average accuracy for Signal-to-Noise Ratios (SNRs)$>$0\,dB, and on the hardware Trojan (HT)-based CC dataset it reaches 94.2\% accuracy, both with fewer than 10k parameters and up to speed of 92 $\mu$s per frame on FPGA/CPU co-execution, surpassing state-of-the-art edge models at a fraction of their cost. These results establish E-SpecFormer as an edge-efficient solution for real-time spectrum intelligence on Internet of Things (IoT) devices. GitHub link to the repository: \url{https://github.com/zsniko/E-SpecFormer}.
\end{abstract}


\section{Introduction}
\label{sec:intro}
\IEEEPARstart{R}{adio} frequency (RF) spectrum monitoring is essential for modern wireless communication systems. Herein, we focus on two specific functions of RF spectrum monitoring, namely automatic modulation recognition (AMR), which is essential for optimizing spectrum usage amid uneven band occupation~\cite{Emad21} with the expansion of 5G/6G and the Internet of Things (IoT), and covert channel (CCs) detection, i.e., malicious signals that subtly exfiltrate sensitive data~\cite{HOST24}, which is essential for defense and surveillance systems. Although AMR and CC detection differ in objectives, both operate on a common input: I/Q samples produced by incoming RF signals down-converted and digitized by the analog-to-digital converters (ADCs) at the RF receiver, providing a digital representation of the RF signal information content.

Traditional likelihood and hand-crafted feature-based methods are computationally expensive or have lower accuracy as modern communication systems become increasingly complex; deep learning has shown strong potential and is dominating the current approaches for AMR~\cite{AMR_benchmark}. However, cloud-based AI inference is unsuitable: transmitting raw RF data to a remote server introduces significant latency, increases communication overhead, and raises data privacy concerns.
Existing AI-based AMR approaches include convolutional neural networks (CNNs) \cite{Emad21, RML18}, long short term memory (LSTM)~\cite{lstm2}, spiking neural networks (SNNs)~\cite{SNN}, hybrid CNN–LSTM~\cite{CLDNN} and CNN–GRU~\cite{CGDNet} models, and more recently Transformers~\cite{MR-Transformer}.
While accurate, these models often contain 100k–20M parameters, with Transformer variants further limited by the quadratic cost of self-attention~\cite{Attention}, making them impractical for edge devices. 
Lightweight alternatives such as Attention-CNN~\cite{Song25} and LSTM-Autoencoder~\cite{DAE} reduce parameter counts below 20k, but typically cover only 11 of the 24 realistic over-the-air modulation schemes in the RadioML2018 dataset~\cite{RML18}. 
Many prior works also rely on input preprocessing (e.g., amplitude/phase normalization, constellation conversion), which requires buffering full I/Q frames and adds latency. Attention could be reduced to linear complexity, enabling much faster inference for long sequences~\cite{sima}. In this work, we focus on end-to-end models that process raw I/Q samples directly for real-time spectrum monitoring.

The hardware Trojan (HT)-based CC dataset~\cite{HOST24} captures realistic CC examples under diverse SNRs and channel impairments. Despite its importance, no hardware accelerator has yet been proposed to address this task.

Although efficient Transformer architectures have been well explored across other domains~\cite{airformer,mb-taylorformer}, a hardware-aware design tailored for RF signal recognition is absent to the best of our knowledge; existing Transformer-based RF approaches require hardware resynthesis for different tasks and exhibit relatively high latency~\cite{MR-Transformer}.

\begin{figure*}[htbp]
\centerline{\includegraphics[width=\textwidth]{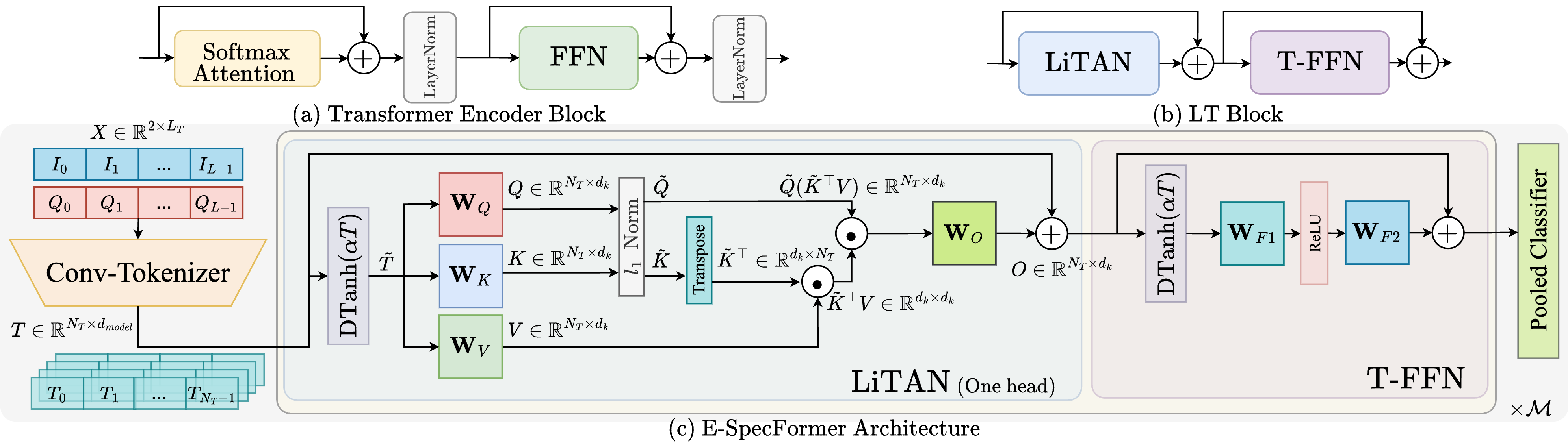}}
\vspace{-1.1em}
\caption{(a) Standard Transformer encoder with quadratic-complexity Softmax attention, LayerNorm, and Feed-Forward Network (FFN) with expansion ratio $r{=}4$. 
(b) Proposed LT (LiTAN + T-FFN) block with DTanh pre-normalization, linearized attention, and Tanh-FFN using $r{=}1$ for Nano/Small (minimizing footprint for extreme edge computing) or $r{=}2$ for Medium/Large (expanding capacity similar to MobileViT-XXS). 
(c) E-SpecFormer, consisting of a Conv-Tokenizer followed by $\mathcal{M}$ stacked LT blocks with scaling depth $\mathcal{M}\!\in\!\{1,2,4,6\}$ for N,S,M, and L variants. $d_k{=}d_{model}{=}d_v=32$ (48 for L only).}

\label{fig:arch}\vspace{-0.4cm}
\end{figure*}

In this paper, we make the following contributions:

\begin{itemize}
    \item We propose \emph{E-SpecFormer}, an edge-efficient Transformer family for end-to-end AMR and CC classification. It is a hybrid network integrating a convolutional tokenizer with \emph{LiTAN} (Linear Tanh Attention Network), a natively hardware-friendly, Softmax- and LayerNorm-free attention mechanism reducing complexity to linear while maintaining or improving classification accuracy.
    
    \item \emph{E-SpecFormer} is parameterized in four sizes: Nano (N), Small (S), Medium (M), Large (L) for hardware adaptation, facilitating deployment from IoT devices to high-performance edge accelerators.
    
    \item A streaming-inference FPGA accelerator for \emph{E-SpecFormer} supporting any input sequence length for the two RF signal recognition tasks without resynthesis.
    
    \item E-SpecFormer pushes the Pareto curve in terms of accuracy vs. computational complexity, outperforming prior art on AMR and CC detection at the edge.
\end{itemize}
\vspace{-0.2cm}

\section{Methodology}
\label{sec:method}

\subsection{Proposed AI Model}
Let $X \in \mathbb{R}^{2 \times L_T}$ denote the input I/Q sequence of token length $L_T$, where each I/Q pair is sampled by the RF front-end. Unlike natural language processing tasks, where tokens are high-dimensional embeddings,  each I/Q token here is only 2D, while the input sequences are long ($L_T$=1024 for RadioML2018 and 640 for HT-CC). In addition to quadratic attention complexity, both Softmax and LayerNorm are major throughput and latency bottlenecks in Transformers, imposing significant overhead on edge devices.  The architecture of the proposed model is depicted in Fig.~\ref{fig:arch}.

\paragraph*{Conv-Tokenizer.}
To address both low token dimensionality and long sequence length, we introduce a convolutional tokenizer (\emph{ConvT}) with kernel size $(2,k)$ and stride $(1,s)$, spanning the I/Q axis over the temporal dimension.  

Non-overlapping patches ($s=k$) are applied for the N and S variants for higher processing efficiency, while the M and L variants employ partly overlapping patches ($s=k/2$) to enrich representations. Each patch is then projected into a $d_{\text{model}}$-dimensional token embedding:
\vspace{-0.1cm}
\[
X \in \mathbb{R}^{2 \times L_T} 
\;\;\xrightarrow{\;\text{Conv}(2,k,s)\;}\;\; 
T \in \mathbb{R}^{N_T \times d_{\text{model}}}, 
 N_T = \Big\lfloor \tfrac{L_T - k}{s} \Big\rfloor + 1.
\]

This operation is mathematically equivalent to a learned projection of flattened patches, yielding higher-dimensional tokens with controllable granularity and efficiency.

\paragraph*{LiTAN}

We introduce LiTAN, integrating:  
(i) a lightweight \emph{dynamic hypertangent} activation as a \emph{pre-norm} layer in place of LayerNorm, and  
(ii) a linearized attention. 

\paragraph*{Pre-Normalization}

Traditional Transformers rely on post-normalization that requires mean-variance statistics, which are costly on edge devices. Recent studies show that explicit normalization is not always required~\cite{DyT}. We leverage a lightweight (\emph{DTanh}) as a hardware-friendly alternative, providing bounded activations and tanh-like scaling without mean–variance statistics:\vspace{-0.1cm}
\[
  \tilde{T}=\sigma_{\alpha,\omega,\beta}(T) \;=\; \tanh(\alpha T)\odot \omega + \beta, 
\]
where $\alpha \in \mathbb{R}$ controls dynamic range, and $(\omega,\beta)$ are per-channel affine parameters broadcast across the sequence. This non-expansive mapping stabilizes scale while avoiding the reductions required by LayerNorm and enables LUT-based quantization making it hardware-friendly.

\paragraph*{Linearized Attention.}

Given $N_T$ tokens with embedding size $d_{model}$, we compute Query, Key, and Value matrices: $Q = \tilde{T} W_Q, K = \tilde{T} W_K,  V = \tilde{T} W_V,$
where weights $W_Q, W_K, W_V \in \mathbb{R}^{d_{\text{model}} \times d_x}$, and typically $d_x=d_{\text{model}}$.

The canonical self-attention mechanism~\cite{Attention} is defined as

\[
\mathrm{Attn}(Q,K,V) = \mathrm{Softmax}\!\left(\frac{QK^\top}{\sqrt{d_k}}\right)V.
\]
Forming the similarity matrix $QK^\top \in \mathbb{R}^{N_T \times N_T}$ entails $\mathcal{O}(N_T^2 d_k)$ operations. Multiplication with $V$ yields a total complexity $\mathcal{O}(N_T^2 d_k)$, which is quadratic in $N_T$.

The linearized attention is defined as:
\[
\mathrm{LinAttn}(Q,K,V) = \tilde{Q} \big( \tilde{K}^\top V \big).
\]
\noindent where $\tilde{Q}=\phi(Q)$, $\tilde{K}=\phi(K)$, and $\phi(\cdot)$ denotes the $l_1$-norm kernel. The computation proceeds in two steps:
\vspace{-0.1cm}
\begin{align*}
S &= \tilde{K}^\top V \quad \in \mathbb{R}^{d_k \times d_v}, \\
\mathrm{LinAttn}(Q,S) &= \tilde{Q} S \quad \in \mathbb{R}^{N_T \times d_v}.
\end{align*}

Both steps cost $\mathcal{O}(N_T d_k d_v)$, yielding an overall linear complexity $\mathcal{O}(N_T d_k d_v)$ with respect to sequence length $N_T$.

The $l_1$-norm scaling of $Q$ and $K$ provides a crucial architectural intuition for RF data. Since I/Q samples exhibit bounded energy and sparse spectral structure, $l_1$ provides a stable magnitude measure without squaring or square-root operations. Intuitively, this improves robustness to channel impairments without requiring costly normalization. Moreover, $l_1$ eliminates costly exponential and cosine operations ($\approx$4x MACs), and integrates naturally with linear attention, which supports streaming inference via associated KV accumulation with dedicated hardware.

\begin{figure*}[htbp]
\centerline{\includegraphics[width=\textwidth]{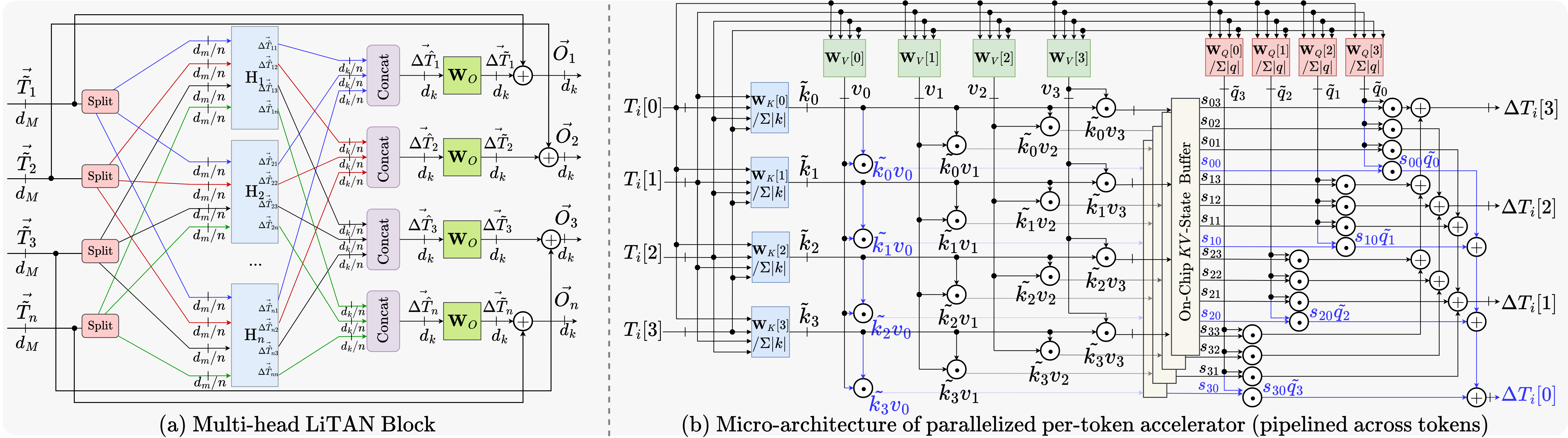}}
\vspace{-1em}
\caption{The proposed LiTAN co-processor. We propose a "Time-Multiplexing" architecture that utilizes spatial parallelism for each token across heads with dedicated compute units ($H_1$…$H_n$) to process their respective vector segments. Across the temporal dimension, the hardware is pipelined: the single input port is time-multiplexed (different colors) to stream tokens ($\vec{T_1}$,$\vec{T_2}$…) cycle-by-cycle, allowing the same hardware to process an arbitrary sequence length.}
\label{fig:litan_hw}\vspace{-0.4cm}
\end{figure*}

A single-head LiTAN block is expressed as:
\[
\mathrm{LiTAN}(T) = T + \mathrm{LinAttn}(	\tilde{T} W_Q, 	\tilde{T} W_K, \tilde{T}W_V)W_o,
\]
In practice, we adopt multi-head LiTAN ($H$=8) (Fig.~\ref{fig:litan_hw}(a)), where concatenating complementary heads enriches representation diversity.
Ultimately, E-SpecFormer reduces complexity from $\mathcal{O}(N_T^2)$ to $\mathcal{O}(N_T)$, making long-sequence I/Q processing feasible on edge devices.

\vspace{-0.3cm}

\begin{table*}[t]
\setlength{\tabcolsep}{5pt}
\centering
\footnotesize
\caption{Comparison of models on RadioML2018 and HT-CC. Average accuracy for SNRs $>$ 0 $dB$. All models are trained on NVIDIA A100 and inference latencies are all evaluated on ARM Cortex-A76 CPU (typical IoT/Embedded) with ONNX Runtime. *Quantized.} \vspace{-0.3cm}
\begin{tabular}{l|cccc|cccc}
\hline
\multirow{2}{*}{\textbf{Model}} &
\multicolumn{4}{c|}{\textbf{RadioML2018}} &
\multicolumn{4}{c}{\textbf{HT-CC}} \\
 & \textbf{Accuracy} & \textbf{\#Param.} & \textbf{kMACs} & \textbf{Latency (ms)} 
 & \textbf{Accuracy} & \textbf{\#Param.} & \textbf{kMACs} & \textbf{Latency (ms)} \\
\hline
CNN (IQ)~\cite{Emad21,HOST24} & 63.3\% & 295,484 & 3,648 & 0.40            & 91.9\% & 184,265 & 2,265 & 0.25\\
Attention-CNN~\cite{Song25} & 63.5\% & 20,390 & 347 & 0.15 & 88.8\% & 19,760 & 345 & 0.14 \\
MobileNetv3-S~\cite{mobilenetv3} & 94.2\% & 1,495,864 & 37,576 & 5.82    & 96.1\% & 1,476,389 & 23,871 & 4.49 \\
CGDNet~\cite{CGDNet} & 66.7\% & 665,874 & 171,545 & 11.30            & 72.2\% & 660,991 & 130,298 & 8.66 \\
CLDNN2~\cite{CLDNN} & 72.8\% & 698,520 & 504,682 & 25.60            & 94.1\% & 619,269 & 314,916 & 16.06  \\
LSTM (IQ)~\cite{lstm2} & 94.2\% & 202,752 & 207,097 & 35.38               & 88.8\% & 200,320 & 129,434 & 22.43 \\
Transformer (4L)~\cite{Attention} & 72.3\% & 35,064 & 336,889 & 156.48     & 86.9\% & 34,437 & 139,776 & 50.11  \\
Performer-N~\cite{performer} & 76.3\% & 8312 & 488 & 1.13 & 91.5\% & 7685 & 305 & 1.01 \\
Linformer-N~\cite{linformer} & 84.0\% & 16054 & 488 & 0.53 & 91.6\% & 15877 & 305 & 0.35 \\
\rowcolor{gray!20} E-SpecFormer-N  \ (This Work) & 86.5\% / 86.3\%* & \textbf{8,314} & \textbf{501} & \textbf{0.33}        & 94.2\% / 94.3\%* & \textbf{7,687} & \textbf{312} & \textbf{0.24}  \\
\rowcolor{gray!20} E-SpecFormer-S  \  (This Work)& 89.4\% / 89.0\%* & 14,780 & 924 & 0.52           & 95.7\% / 95.6\%* & 14,153 & 577 & 0.38 \\
\rowcolor{gray!20} E-SpecFormer-M (This Work)& 92.7\% / 92.5\%* & 36,032 & 4,589 & 1.55              & 96.1\% / 96.0\%* & 35,405 & 2,854 & 1.04 \\
\rowcolor{gray!20} E-SpecFormer-L  \ (This Work)& \textbf{94.0\% / 93.6\%*} & 116,532 & 15,040 & 3.06    & \textbf{96.9\% / 96.8\%*} & 115,601 & 9,355 &  2.04 \\
\hline
\end{tabular}
\label{tab:comparison}\vspace{-0.5cm}
\end{table*}

\subsection{Proposed Hardware Accelerator}

The ConvT IP is implemented in a streaming mode, using stationary weight kernels to process I/Q samples from ADC on-the-fly without buffering, following the style of~\cite{Song25}.

The proposed hardware accelerator architecture for LiTAN exploits the algebraic properties of LinAttn to decouple hardware resources from sequence length. As illustrated in Fig.~\ref{fig:litan_hw}, the design employs a dual-parallelism strategy: fine-grained spatial parallelism within each token processing step and coarse-grained temporal pipelining across the token stream.

To maximize throughput, the computation for a single time step $t$ is spatially unrolled. The input embedding vector $T_t$ is broadcast to a parallel array of $H$ independent head compute units, corresponding to multi-head attention where the input is split into $d_{\text{model}}/H$ segments. Inside each unit, the accelerator executes projection and state-update operations concurrently. The input segment is projected into Query ($\tilde{q}$), Key ($\tilde{k}$), and Value ($v$) vectors and the $l_1$ norm has the advantage of being computed on-the-fly per token with adder tree and reciprocal approximation. The KV operation is done using parallel multiplier banks. Crucially, the recurrent state update $S_{t} \leftarrow S_{t-1} + \tilde{k}^\top v$ and the compensation vector $\Delta T_t \leftarrow S_{t} \tilde{q}$ are computed in a strictly local dataflow. Our design performs streaming inference with similarity kernel decomposition and compression into a recurrent state matrix :
\vspace{-0.15cm}
\[
S_t = \Sigma_{i=1}^{t} \phi (k_i)^\top v_i
\]
\vspace{-0.1cm}
The on-chip KV-state remains constant with respect to sequence length, maintaining a fixed memory footprint as $T$ increases. This stands in sharp contrast to canonical attention, which requires storing full history and computing an $\mathcal{O}(T^2)$ attention matrix, creating bandwidth and storage bottlenecks that typically dominate Transformer hardware. While optimization schemes like FlashAttention mitigate these bottlenecks via SRAM tiling, our design eliminates the quadratic memory dependency entirely, leveraging locality without requiring sequence blocking.

The physical single input port is time-multiplexed; tokens arrive serially and flow through the spatial fabric described above. We achieve an initiation interval of 1: the pipeline accepts a new token every clock cycle while previous tokens propagate through the deeper stages of the datapath. Consequently, the hardware implementation remains invariant regardless of input length. Scaling $T$ results only in a linear increase in total execution time, with no degradation in per-token latency or resource utilization.

\vspace{-0.1cm}
\section{Results and Discussion}
\label{sec:results}

\begin{figure}[t]
\begin{minipage}[b]{1.0\linewidth}
  \centering
  \centerline{\includegraphics[width=9.5cm]{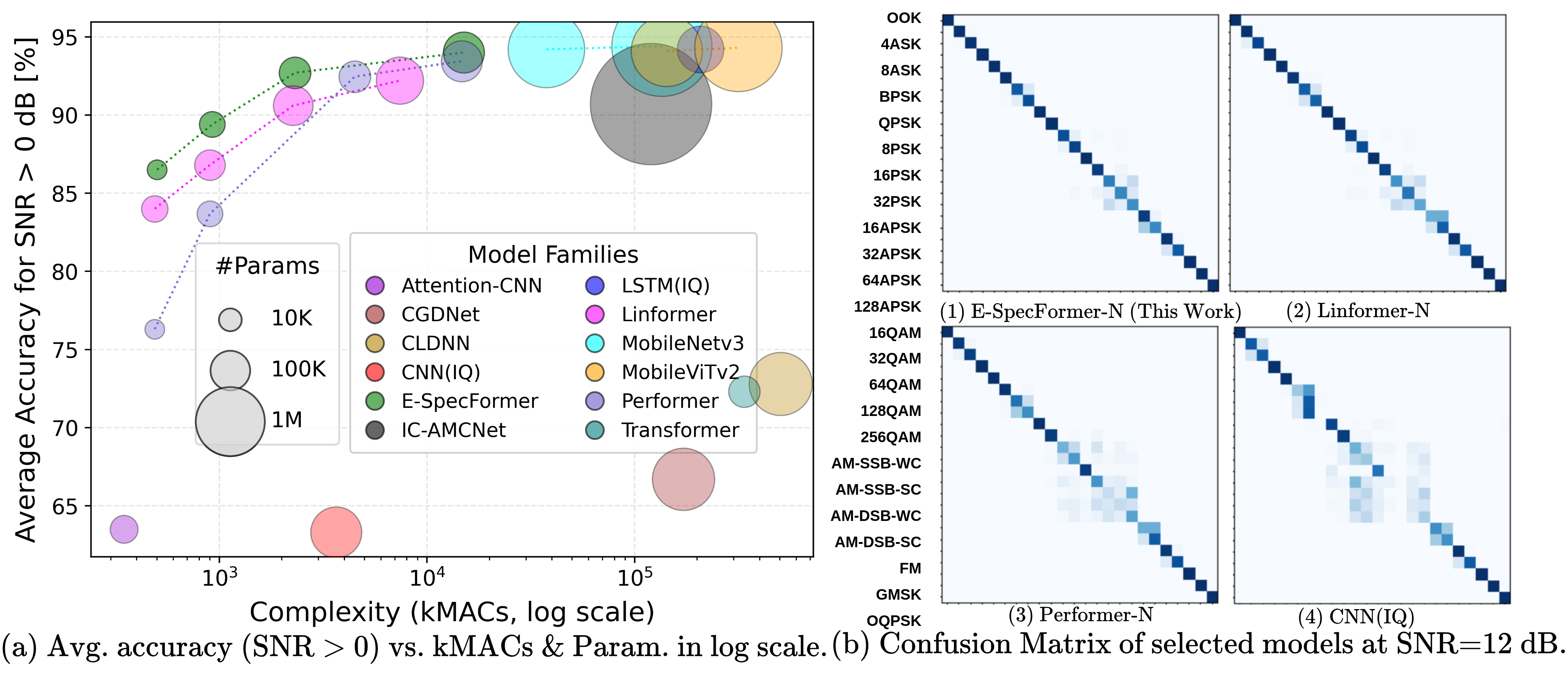}}
\end{minipage}
\vspace{-2.5em}
\caption{Detailed AMR analysis. All confusion matrices share identical axis ordering (OOK to OQPSK). E-SpecFormer-N demonstrates superior differentiation of overlapping higher-order schemes compared to baselines.}
\label{fig:complexity}\vspace{-0.35cm}
\end{figure}

We evaluate the proposed models on the RadioML2018 ($L_T\!=\!1024$) and HT-based CC ($L_T\!=\!640$) datasets. To rigorously isolate our architectural contributions, we baseline against RF-specific models, and tightly matched efficient linear attention Transformers (Performer~\cite{performer}, Linformer~\cite{linformer}) equipped with the same conv tokenizer and identical scaling. Due to space constraints, Table~\ref{tab:comparison} lists the Nano variants, while Fig.~\ref{fig:complexity} plots all variants. All reported models are trained using Adam (lr=$10^{-3}$) on an NVIDIA A100 GPU (Intel Xeon 6300 CPU + 128GB RAM). Inference metrics (accuracy, parameters, kMACs, latency) are assessed on an ARM Cortex-A76 CPU (typical IoT / edge device) via ONNX Runtime. This IoT setup ensures a hardware-agnostic fair comparison.

\begin{figure}[t]
\begin{minipage}[b]{1.0\linewidth}
  \centering
  \centerline{\includegraphics[width=9cm]{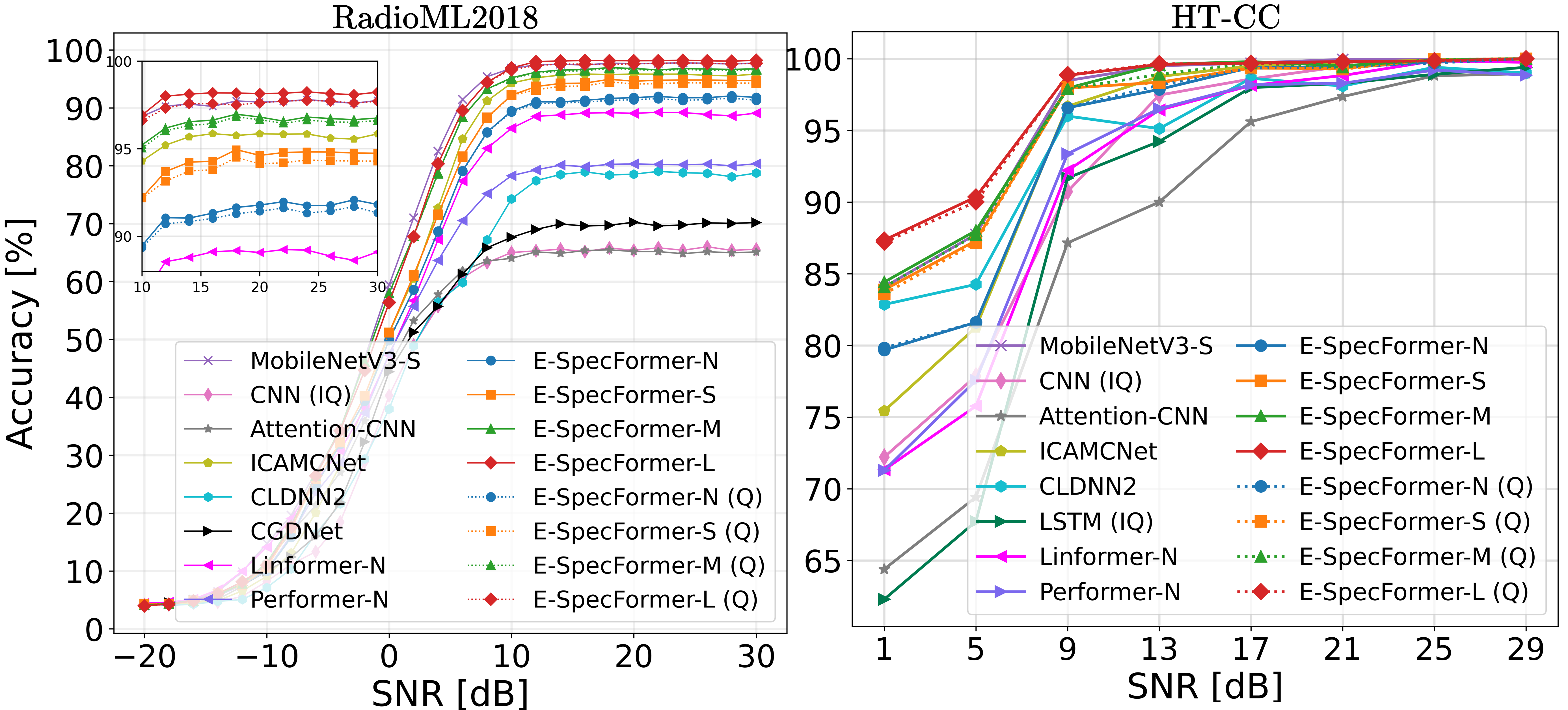}}
\end{minipage}
\vspace{-2.2em}
\caption{Accuracy vs. SNR curves. Dotted lines with (Q) indicate post-training INT8 quantized variants with consistently low accuracy drop across SNRs.}
\label{fig:acc_snr}\vspace{-0.35cm}
\end{figure}

\paragraph*{Results on RadioML2018}
Table~\ref{tab:comparison} summarizes performance. Existing lightweight AMR models (typically $<$65\% accuracy) target small-scale 11-class tasks, while MobileNetv3-S achieves 94.2\% but requires 1.5M parameters. E-SpecFormer offers superior efficiency-accuracy scaling. The N variant attains 86.5\% accuracy with 8.3k parameters and 330 $\mu$s latency, drastically outperforming edge baselines. The L variant achieves 94.0\%, matching MobileNetv3-S with 92\% fewer parameters. Accuracy–SNR curves (Fig.~\ref{fig:acc_snr}(a)) show that all variants remain competitive across SNRs, with the L variant exceeding all baselines at high SNRs.
\begin{table}[t]
\centering
\footnotesize
\setlength{\tabcolsep}{5pt}
\caption{Ablation study for N and M variants on RadioML2018 and HT-CC datasets. SNR $>$ 0 $d$B.}
\begin{tabular}{ccc|cc|cc}
\hline
\multicolumn{3}{c|}{\textbf{Modules}} & \multicolumn{2}{c|}{\textbf{RadioML2018}} & \multicolumn{2}{c}{\textbf{HT-CC}} \\
ConvT & DTanh & LinAttn & N & M & N & M \\
\hline
- & - & - & 67.1\% & 72.3\% & 65.3\% & 86.9\% \\ 
\checkmark & - & - & 79.9\% & 92.7\% & 92.9\% & 96.2\% \\ 
\checkmark & \checkmark & - & 79.9\% & 92.6\% & 92.8\% & 96.0\% \\ 
\checkmark & -  &  \checkmark  & 83.0\% & 92.7\% & 93.8\% & 96.2\% \\ 
\rowcolor{gray!20}
\checkmark & \checkmark &  \checkmark & 86.5\% & 92.7\% & 94.2\% & 96.2\% \\
\hline
\end{tabular}

\label{tab:ablation}\vspace{-0.5cm}
\end{table}

\paragraph*{AMR \& RF Waveform Analysis}
To validate the architecture's physical RF relevance, Fig.~\ref{fig:complexity}(b) analyzes 12\,dB SNR confusion matrices. Baseline CNNs and Performer-N suffer RF bottlenecks when processing raw I/Q streams: they conflate 16/32PSK (tightly packed phase states, making time-domain I/Q trajectories similar), higher-order 64/128/256QAM (reliance on fine-grained amplitude/phase variations easily corrupted), and AM-SSB/DSB-WC/SC (carrier presence masked by noise). While Linformer-N improves on PSK, it still conflates dense QAMs. E-SpecFormer better separates these highly correlated sequences, confirming it learns fundamental dynamics directly from raw baseband samples, rather than relying on superficial statistical artifacts.

\paragraph*{Results on HT-CC (Generalization)}
Performance on HT-CC demonstrates the model's robustness to distinct RF tasks and channel distribution shifts. E-SpecFormer consistently surpasses dedicated RF baselines: the N variant (94.2\%) outperforms all prior models, while the L variant achieves up to 96.9\%, setting the highest benchmark. Fig.~\ref{fig:acc_snr}(b) further illustrates this stability across diverse SNRs.

\paragraph*{Ablation \& Complexity–Accuracy Trade-off}
Table~\ref{tab:ablation} isolates module contributions. ConvT provides the primary representational gain, highlighting the importance of temporal–spectral feature extraction and expanding the low-dimensional I/Q tokens into richer embeddings. DTanh stabilizes training and enables LUT-based quantization, directly eliminating the overhead of hardware-unfriendly LayerNorm while preserving accuracy. In extremely shallow models (N), DTanh and linear attention jointly improve accuracy. For larger variants (M), accuracy is already high due to increased model capacity; however, LiTAN reduces kMACs by 35.7/49.6\% (N/M) on RadioML2018 and 25.2/37.5\% on HT-CC against quadratic attention without accuracy loss on the M variant, satisfying the edge-computing goal. Fig.~\ref{fig:complexity}(a) plots accuracy against kMACs and parameters for the resulting Pareto frontier. We evaluate adapted Performer and Linformer baselines under the same scaling. Under extreme edge constraints, matched linear baselines either degrade severely (Performer-N which uses random feature mapping to approximate the Softmax kernel drops to 76.3\%) or bloat (Linformer-N requires nearly double the parameters for 84.0\%, as it introduces two sequence-dependant learned projection matrices). While some CNN-based models remain similarly lightweight, their accuracy plateaus well below 70\%. E-SpecFormer strictly dominates this low-complexity regime, pushing the frontier for real-time edge computing.

\paragraph*{Quantization}
For FPGA deployment, we apply post-training integer quantization: 8-bit weights (memory bandwidth is dominated by weights) and 16-bit activations (W8A16), as INT8 activations can distort low-SNR amplitude, phase and temporal correlations and RF signal features are sensitive to activation quantization noise unlike images with bounded pixels. The model is calibrated offline with MinMax on a representative subset of each training dataset. As detailed in Table~\ref{tab:comparison} and Fig.~\ref{fig:acc_snr}, the accuracy drop is negligible: on RadioML the average drop is $\le$0.4\%, and on HT-CC the quantized models exhibit virtually no degradation.

\paragraph*{Hardware Implementation \& Results}

The proposed design is implemented on a Zynq™ UltraScale+™ MPSoC ZCU104 FPGA with Vivado and Vitis 2025.1. Evaluation is performed by replaying held-out test set frames through the same AXI-Stream interface, replicating the streaming data path under realistic signal conditions. Profiling E-SpecFormer on the on-board Cortex-A53 CPU reveals the bottlenecks: ConvT (150$\mu$s) and LiTAN (578$\mu$s) dominate, whereas FFN (26$\mu$s), and the pooled classifier (5$\mu$s) are highly efficient on CPU due to contiguous memory access and effective use of SIMD and cache locality. ConvT and LiTAN are implemented on programmable logic (PL), designed to stream I/Q samples from the ADC and communicating directly via on-chip AXI-Stream. Crucially, LiTAN utilizes a localized on-chip KV-state buffer that updates recurrently; it never computes or stores an $\mathcal{O}(N_T^2)$ similarity matrix, completely eliminating the DRAM bandwidth bottlenecks of canonical Transformers. FFN computations are offloaded to CPU, ensuring flexibility across varying expansion ratios that changes neuron numbers yet remain easily handled in software. CPU-PL communication overhead is highly optimized: e.g., for the N variant, following ConvT downsampling ($N_T\!=\!64$, $d_{model}\!=\!32$), the LiTAN accelerator returns a 4\,KB 16-bit activation payload via DMA. Including CPU cache synchronization, this transfer overhead is $<5\,\mu$s. For deeper variants, the DMA efficiently transfers these payloads between software FFNs and the hardware LiTAN IP. The LiTAN block (Fig.~\ref{fig:litan_hw}) is fully reusable across variants, enabling direct scalability without hardware redesign. The end-to-end latency of 124 $\mu$s includes DMA/AXI and comprises ConvT-PL (25 $\mu$s) and LiTAN-PL (51 $\mu$s), with Dense layers' latency same on CPU. It is 6.3x faster than the A53 CPU with only 0.72W power overhead. Even including on-board CPU power, the heterogeneous design achieves 40\% of the speed of an RTX 5070Ti while being 42× more power efficient.

The architecture scales cleanly across model sizes. Measured latencies for the N/S/M/L variants are: AMR 124/243/481/719$\mu$s, HT-CC 92/181/359/537$\mu$s. Inter-token pipelining makes the design invariant to input sequence length, overcoming a key limitation of MR-Transformer~\cite{MR-Transformer}, which requires re-synthesis and place-and-route upon changing the dataset/task while exhibiting higher latency (2ms). Our single bitstream supports arbitrary $T$ without modification. 

Compared with recent state-of-the-art FPGA and GPU implementations (Table~\ref{tab:hw_comparison}), the proposed heterogeneous accelerator offers the lowest latency–power product as our figure of merit. Spike-based designs~\cite{SNN} achieve lower power but target only 11-class AMR and yield reduced accuracy, while the GPU approach of~\cite{tcas2amr} achieves high speed at $\sim$120~W power, unsuitable for embedded RF applications. In contrast, our design provides a Pareto-optimal balance of accuracy, latency, and power on realistic edge hardware.

\begin{table}[!t]
\centering
\footnotesize
\caption{Comparison with recent SOTA GPU and FPGA implementations on AMR. We tested inference of our model on RTX 5070 Ti GPU (Different from the A100 training GPU. Float16 \& TensorRT): 49.6$\mu$s (N), 70.9$\mu$s (S), 127.2$\mu$s (M), 237.0$\mu$s (L). *N variant}
\label{tab:hw_comparison} \vspace{-0.2cm}
\renewcommand{\arraystretch}{1.2}
\setlength{\tabcolsep}{1pt}
\begin{tabular}{l l c c c c c c}
\toprule
\textbf{Work} & \textbf{Platform} & \textbf{Quant.} & \textbf{LUT} & \textbf{FF} & \textbf{DSP} & \textbf{(PL)Power} & \textbf{Latency} \\
\midrule
TCAS-II'22~\cite{tcas2amr} & GTX1660Ti & 32-bit & -- & -- & -- & $\sim$120W & 82.9$\mu$s  \\
\textbf{This Work} & \textbf{RTX5070Ti} & \textbf{16-bit} & -- & -- & -- & \textbf{$\sim$140W} & \textbf{49.6$\mu$s*}  \\
\midrule
TAI'24~\cite{SNN} & PYNQ & 16-bit & 31k & 51k & 0 & 1.67W & 176.8$\mu$s \\
TWC'25~\cite{MR-Transformer} & XCZU3EG & 16-bit & 43k & 29k & 328 & 0.8W & 2 ms  \\
\textbf{This Work} & \textbf{ZCU104} & \textbf{8-bit} & \textbf{73k} & \textbf{48k} & \textbf{832} & \textbf{0.72W} & \textbf{92/124$\mu$s*}  \\
\bottomrule
\end{tabular}\vspace{-0.3cm}
\end{table}


\section{Conclusion}
We introduced E-SpecFormer, a super-lightweight Transformer family for RF spectrum monitoring on edge devices. At the core of our design is LiTAN, a Softmax- and LayerNorm-free attention block enabling efficient long-sequence I/Q processing in linear time with intra-token parallelization and inter-token pipelining, enabling sequence length adaptation without hardware redesign. Across two benchmark datasets, E-SpecFormer achieves state-of-the-art accuracy while reducing parameters and complexity by orders of magnitude, with the N variant delivering real-time throughput of sub-100$\mu$s with fewer than 10k parameters, establishing E-SpecFormer as a practical solution for spectrum intelligence at the edge. 
\bibliographystyle{ieeetr}
\bibliography{refs}



\end{document}